\documentclass[lettersize,journal]{IEEEtran}
\usepackage{amsmath,amsfonts}
\usepackage{algorithmic}
\usepackage{algorithm}
\usepackage{array}
\usepackage[caption=false,font=normalsize,labelfont=sf,textfont=sf]{subfig}
\usepackage{textcomp}
\usepackage{stfloats}
\usepackage{url}
\usepackage{verbatim}
\usepackage{graphicx}
\usepackage{cite}
\usepackage{booktabs}
\usepackage{multirow}
\usepackage{makecell}
\usepackage{array}
\usepackage{tabularx}
\begin{document}

\title{Beyond Modeling Order: Dual-Path Spatiotemporal Representation Learning for Cross-Subject EEG Emotion Recognition}

\author{
	Zehui Xiao,
	Yi Zheng,
	Guixing Fu,
	Wenkai Lu,
	Lingfei He,
	Tao Tan,
	Yue Liu,
	Silun Wang,
	Junhai Xu,
	and Mengting Liu%
	\thanks{Zehui Xiao and Yi Zheng contributed equally to this work.}%
	\thanks{Zehui Xiao, Yi Zheng, Guixing Fu, and Mengting Liu are with the School of Biomedical Engineering, Shenzhen Campus of Sun Yat-sen University, Shenzhen 518107, China (e-mail: xiaozh26@mail2.sysu.edu.cn; zhengy323@mail2.sysu.edu.cn; fugx3@mail2.sysu.edu.cn; liumt55@mail.sysu.edu.cn).}%
	\thanks{Wenkai Lu, Lingfei He, and Junhai Xu are with the School of Computer Science and Technology, Tianjin University, Tianjin 300354, China (e-mail: 3020244261@tju.edu.cn; lingfei@tju.edu.cn; jhxu@tju.edu.cn). Tao Tan and Yue Liu are with the Faculty of Applied Sciences, Macao Polytechnic University, Macao, China (e-mail: taotan@mpu.edu.mo; yue.liu@mpu.edu.mo). Silun Wang is with Shenzhen Dr. Brain Technologies Co., Ltd., Shenzhen 518057, China (e-mail: dr.brain@drbrain.com.cn).}%
	\thanks{Junhai Xu and Mengting Liu are the corresponding authors (e-mail: jhxu@tju.edu.cn; liumt55@mail.sysu.edu.cn).}%
	\thanks{This work was supported by the Joint Funds of the National Natural Science Foundation of China under Grant U24A20340, the Shenzhen Science and Technology Innovation Program under Grant JCYJ20240813112459030, the National Natural Science Foundation of China under Grant 62671639, and the Science and Technology Development Fund, Macau SAR (File No. 0170/2026/AGJ) under the FDCT-GDST Joint Funding Scheme.}%
}

\maketitle

\begin{abstract}
Cross-subject generalization remains a major challenge for EEG-based emotion recognition due to pronounced inter-subject variability in neural activity and emotion-related spatiotemporal patterns. Existing methods mainly alleviate inter-subject differences through distribution alignment or contrastive learning, while paying limited attention to representation bias associated with fixed spatiotemporal modeling orders. We hypothesize that high-level emotion representations of the same EEG signal should remain consistent across different spatiotemporal modeling paths. Based on this hypothesis, we propose a Dual-Path Spatiotemporal Contrastive Learning (DP-STCL) framework for cross-subject EEG emotion recognition.DP-STCL employs two complementary encoders with independent parameters, namely a Temporal-to-Spatial (TS) path and a Spatial-to-Temporal (ST) path, to model EEG spatiotemporal information under different integration orders. During representation learning, the same emotional stimulus is used to establish cross-subject associations, while three complementary relations jointly constrain representation variations across subjects and modeling paths. This promotes stable and discriminative stimulus-related emotion representations. To further handle limited labeled target-subject data, we introduce a LoRA-based few-shot adaptation strategy with low parameter overhead. Extensive cross-subject experiments on SEED, SEED-IV, SEED-V, and FACED show that DP-STCL consistently outperforms the compared methods, demonstrating its effectiveness in improving cross-subject EEG emotion recognition.
\end{abstract}


\section{Introduction}
\label{sec:introduction}

\IEEEPARstart{E}{lectroencephalography} EEG-based emotion recognition has attracted considerable attention in affective computing and brain-computer interfaces, with applications in mental health assessment, human-computer interaction, and intelligent healthcare \cite{ref1,ref2,ref3,ref4,ref5}. In recent years, deep learning methods have significantly improved the performance of EEG-based emotion recognition \cite{ref6,ref7,ref8,ref9,ref10}. However, these models often rely heavily on the training subjects and show substantial performance degradation when applied to unseen subjects. This performance degradation is largely caused by substantial inter-individual variability in EEG signals. Different subjects exhibit considerable differences in neural activity patterns, spatial distributions, signal statistics, and noise levels \cite{ref11,ref12,ref13}. Therefore, improving generalization to unseen subjects is critical for practical EEG emotion recognition.
To address this issue, existing research has primarily focused on reducing inter-subject distribution differences, employing methods such as domain adaptation, transfer learning, and domain-adversarial learning to mitigate statistical shifts between subjects through feature distribution matching or domain-invariant representation learning \cite{ref12,ref14,ref15,ref16,ref37,ref40}. In recent years, contrastive learning has also been applied to EEG representation learning by constructing positive relations across different views or subjects to encourage the learning of shared representations \cite{ref13,ref22,ref23,ref24,xie2026ta2cl}. Although these methods improve cross-subject generalization,they mainly focus on reducing subject-related variation while paying
limited attention to how EEG spatiotemporal information itself is modeled.

Emotion-related EEG activity inherently exhibits complex spatiotemporal coupling: spatial coordination patterns among brain regions evolve dynamically over time, while local temporal dynamics can in turn shape and reorganize spatial interactions across regions \cite{liu2021context,liu2021fleeting}. Inter-subject variability further increases the diversity of these spatiotemporal patterns, making them difficult to characterize using a single fixed modeling scheme \cite{ref11,ref12,ref13}. 
However, existing methods typically adopt a predefined spatiotemporal modeling order, such as spatial-to-temporal or temporal-to-spatial processing \cite{ref35,ref51}.
Because temporal and spatial information are strongly coupled, applying these operations in different orders may emphasize different aspects of the same EEG signal. A single modeling order may therefore introduce a structural preference toward the features emphasized by that path and overlook complementary spatiotemporal patterns. Furthermore, for the same EEG segment, different spatiotemporal modeling paths may capture local features with different emphases, but their high-level emotion-related representations should remain consistent or compatible. In other words, if the model truly learns stable stimulus-related emotion representations rather than modeling-path bias introduced by the information integration order, the representations produced by different modeling paths should ultimately converge toward a shared high-level semantic space.

Based on these considerations, we propose a Dual-Path Spatiotemporal Contrastive Learning (DP-STCL) framework for cross-subject EEG emotion recognition. DP-STCL employs two complementary encoders with independent parameters: a Temporal-to-Spatial (TS) path and a Spatial-to-Temporal (ST) path.
The TS branch first applies temporal convolution to EEG signals to capture multiscale temporal dynamics and then models dependencies across brain regions through spatial convolution. In contrast, the ST branch first applies spatial convolution to aggregate whole-brain spatial information and then models its temporal evolution through temporal convolution. Through this dual-path design, the model learns EEG spatiotemporal features from different perspectives, thereby enhancing representation diversity and robustness.
During representation learning, the same emotional stimulus is used as the basis for establishing cross-subject relations, and three complementary relations are constructed to account for subject variation and differences between the two modeling paths. The same-subject cross-path relation enhances the consistency between representations of the same EEG sample obtained through the TS and ST modeling paths, thereby reducing modeling path bias introduced by different information integration orders. The same-path cross-subject relation encourages different subjects exposed to the same emotional stimulus to form similar representations, thereby reducing subject-specific variations. The cross-subject cross-path relation further promotes the stability of emotion representations when both the subject and the modeling path vary simultaneously. Through the joint constraints of these three relations, the model learns more stable, subject-invariant, and discriminative emotion representations, thereby improving the generalization ability of cross-subject EEG emotion recognition. 

Furthermore, although subject invariant representations can improve a model's generalization to unseen subjects, individual differences may still persist in practical applications and cannot be fully eliminated through a unified representation. At the same time, acquiring a large amount of labeled EEG data for new subjects is typically costly. Therefore, this paper further addresses individual adaptation when only a small number of labeled samples are available for the target subject. Building upon a frozen pretrained dual-path encoder, we adopt a strategy that combines few-shot classification with Low-Rank Adaptation (LoRA) \cite{ref28,ref29,ref30}. Specifically, we first freeze the parameters of the pretrained encoder and train only the classifier on data from multiple subjects. Subsequently, in the few-shot setting for a target subject, we fine-tune only the LoRA parameters in the classifier, thereby achieving efficient subject adaptation with very few labeled samples. This design significantly reduces training costs and the risk of overfitting in downstream tasks while maintaining the model's generalization performance.

The main contributions of this paper are summarized as follows:

\begin{itemize}
	\item We propose a dual-path spatiotemporal contrastive learning framework for cross-subject EEG emotion recognition. The framework exploits the spatiotemporal structural characteristics of EEG signals through two complementary modeling paths, including a temporal feature driven spatial correlation mining paradigm and a spatial pattern driven temporal dynamics mining paradigm.
	
	\item We use the same emotional stimulus as the basis for establishing cross-subject associations and explicitly model three complementary relations, including same-path cross-subject, same-subject cross-path, and cross-subject cross-path relations. Same-path cross-subject relations are optimized through contrastive learning, whereas same-subject and cross-subject cross-path relations are incorporated into semantic alignment. These complementary relations jointly constrain representation stability under subject variation and modeling-path variation, thereby learning more robust emotion-related representations for cross-subject recognition.
	
	\item We introduce a subject adaptation strategy that combines few-shot learning with LoRA in downstream tasks, enabling efficient cross-subject emotion recognition under limited sample conditions.
	
	\item Extensive experiments on multiple public EEG emotion datasets demonstrate the effectiveness and superiority of the proposed method for cross-subject emotion recognition.
\end{itemize}

\section{Related Work}
\label{sec:related_work}

\subsection{Cross-Subject Emotion Recognition}

Traditional EEG emotion recognition primarily relies on supervised
learning with handcrafted features and conventional classifiers. In
single-subject settings, frequency-domain energy, time-domain features,
and nonlinear indices combined with classifiers such as support vector
machines can achieve good performance \cite{ref31,ref32}. Deep learning
methods have further improved single-subject emotion recognition by
using convolutional neural networks (CNNs) and recurrent neural
networks (RNNs) to capture spatial and temporal patterns
\cite{ref6,ref8,ref35,ref48}. For example, 2D CNNs model regional
relationships from spatial topological maps, while LSTMs capture
dynamic information in time series. Transformer architectures have also
been introduced to model long-range dependencies and improve
multi-emotion recognition. However, in cross-subject scenarios,
differences in EEG statistical distributions and spatial patterns can
cause substantial performance degradation when models are directly
applied to unseen subjects \cite{ref12,ref37}.

Various cross-subject adaptation strategies, including domain adaptation
\cite{ref12}, transfer learning \cite{ref37}, and domain-adversarial
learning \cite{ref40}, have been proposed to mitigate distribution
shifts between subjects. Many methods align feature distributions across
subjects to learn more consistent representations
\cite{ref12,ref14,ref15,ref16,ref37,ref40}. Although these approaches
improve cross-subject generalization, they still pay limited attention
to the spatiotemporal structure and semantic consistency of EEG
signals. This limitation motivates the structured contrastive learning
mechanism proposed in this paper.

\subsection{Contrastive Learning for EEG Representation Learning}

Contrastive learning learns discriminative representations by
constructing positive and negative sample pairs and has achieved
significant success in computer vision \cite{ref20,ref21}. In recent
years, it has also been increasingly applied to EEG representation
learning. Existing methods commonly use data augmentation or temporal
transformations to construct different views of the same signal as
positive pairs \cite{ref22,ref23,ref24,ref20}. Banville et al. learned
EEG representations using self-supervised contrastive objectives and
improved downstream-task performance \cite{ref22}. Jiang et al. applied
self-supervised contrastive learning to EEG-based sleep staging,
further demonstrating its effectiveness in EEG time-series modeling
\cite{ref24}. Shen et al. proposed CLISA \cite{ref13}, which improves
cross-subject emotion recognition by maximizing the similarity between
representations of different subjects exposed to the same emotional
stimulus. More recently, Xie et al. proposed temporal asynchronous
alignment-based contrastive learning to account for inter-subject
temporal misalignment in cross-subject EEG emotion recognition
\cite{xie2026ta2cl}. Overall, these studies demonstrate the
effectiveness of contrastive learning for robust cross-subject EEG
representation learning.

\subsection{Spatiotemporal Feature Modeling and Dual-Path Architectures}

EEG signals exhibit distinct spatiotemporal coupling characteristics,
with emotional information involving both temporal dynamics and spatial
coordination across brain regions. Effectively capturing this structure
is therefore important for EEG representation learning. Previous studies
have typically used CNNs or CNN-RNN architectures to extract local
spatiotemporal features, but direct integration of temporal and spatial
information may limit the modeling of their structural relationships
\cite{ref35,ref48}. Hierarchical temporal structures and attention
mechanisms have been introduced to enhance temporal dependency modeling
\cite{ref51,ref52}. Spatiotemporal attention can dynamically assign
weights according to input signals to improve feature representation
\cite{ref52}, while graph-based neural networks exploit topological
structures to model spatial relationships among brain regions
\cite{ref53,li2026surfgnn,zeng2024adaptive}. However, existing methods
typically adopt fixed spatiotemporal modeling schemes or information
integration orders and rarely represent EEG spatiotemporal structures
from complementary modeling paths. Motivated by this limitation, this
paper draws on multi-view learning \cite{ref27} and models the
Temporal-to-Spatial and Spatial-to-Temporal paths in parallel. The two
paths characterize EEG spatiotemporal structure under different
integration orders and provide complementary representations.
Cross-path consistency constraints further guide the two paths toward
stable representations in a shared semantic space.

\section{Methods}
\label{sec:methods}

\subsection{Overall Framework}
\label{subsec:overall_framework}

\begin{figure*}[!t]
	\centering
	\includegraphics[width=\textwidth]{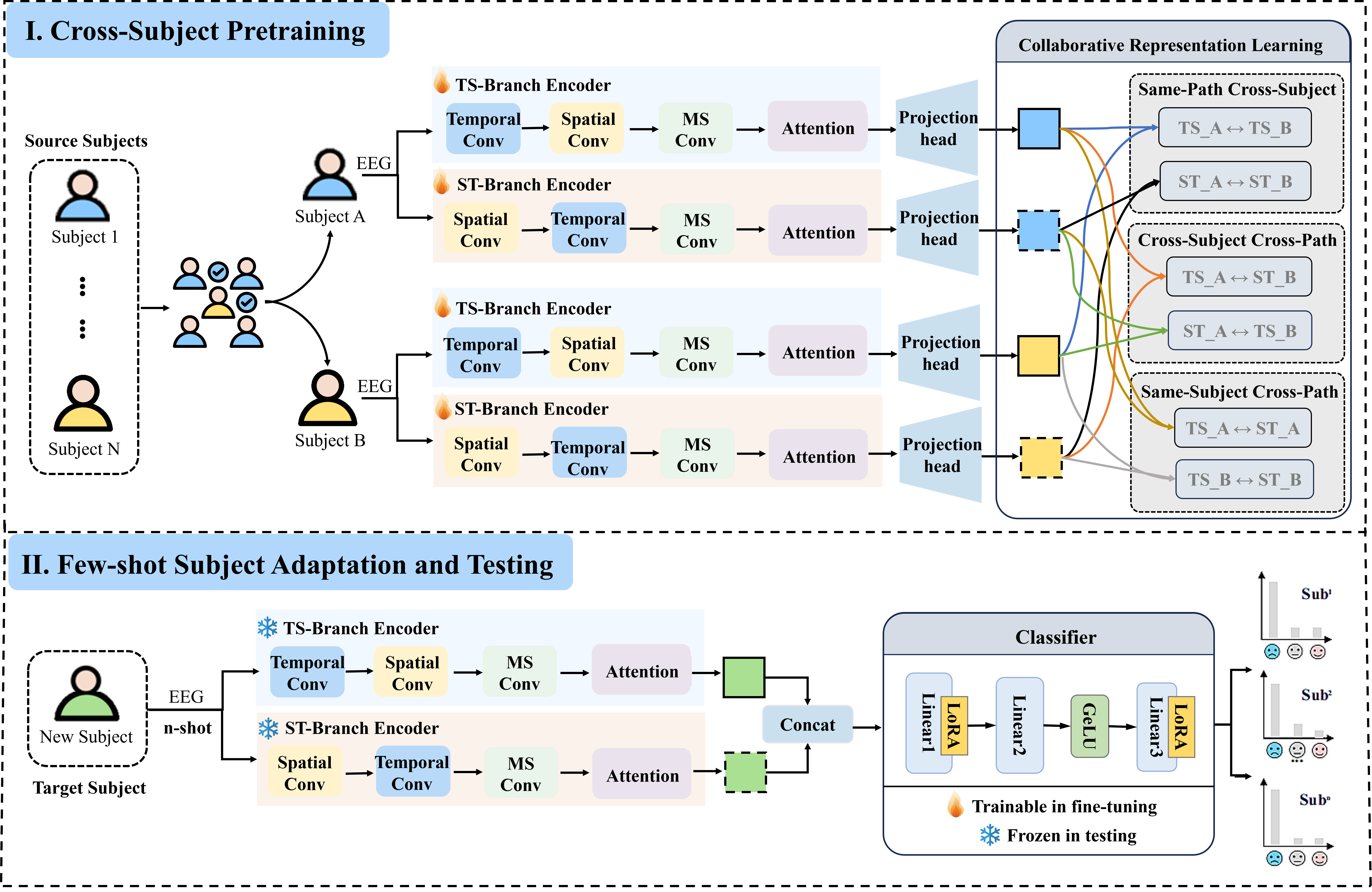}
	\caption{Overall framework of the proposed DP-STCL for cross-subject EEG emotion recognition. The framework consists of cross-subject pretraining with dual-path spatiotemporal encoding, same-path cross-subject contrastive learning, and cross-path semantic alignment, followed by few-shot subject adaptation with LoRA and emotion recognition for the target subject.}
	\label{fig:overall_framework}
\end{figure*}

To address inter-subject variability and representation bias associated with fixed spatiotemporal modeling orders, we propose the Dual-Path Spatiotemporal Contrastive Learning (DP-STCL) framework, as illustrated in Fig.~\ref{fig:overall_framework}. DP-STCL contains two independent encoding paths, namely the Temporal-to-Spatial (TS) path and the Spatial-to-Temporal (ST) path, which model the same EEG signal under different information integration orders. The resulting high-level representations are jointly optimized through same-path cross-subject contrastive learning and cross-path semantic alignment.
Specifically, three complementary relations are constructed to account for subject variation and modeling-path variation, including same-path cross-subject, same-subject cross-path, and cross-subject cross-path relations. After pretraining, the projection heads are removed, and the features extracted by the TS and ST encoders are used for downstream emotion classification. For few-shot adaptation to an unseen target subject, the pretrained encoders are frozen, and Low-Rank Adaptation (LoRA) is introduced into the classifier for parameter-efficient adaptation.

\subsection{Dual-Path Spatiotemporal Feature Encoding}
\label{sec:dual_path_encoding}

 We construct two spatiotemporal encoding paths, $f_{\mathrm{TS}}(\cdot)$ and $f_{\mathrm{ST}}(\cdot)$, with independent parameters. The two paths do not represent actual information propagation directions in brain activity, but model the same spatiotemporally coupled EEG signal using different information integration orders.
\subsubsection{Dual-Path Spatiotemporal Encoding}

Both paths consist of temporal and spatial encoding operations. 
Let $\mathcal{T}^{p}(\cdot)$ and $\mathcal{S}^{p}(\cdot)$ denote the temporal and spatial encoding operations of path $p\in\{\mathrm{TS},\mathrm{ST}\}$, respectively:
\begin{equation}
	\begin{aligned}
		\mathcal{T}^{p}(H)
		&=
		\phi\!\left(
		\mathrm{BN}\!\left(
		\mathrm{Conv}_{t}^{p}(H)
		\right)
		\right), \\
		\mathcal{S}^{p}(H)
		&=
		\phi\!\left(
		\mathrm{BN}\!\left(
		\mathrm{Conv}_{s}^{p}(H)
		\right)
		\right),
	\end{aligned}
	\label{eq:temporal_spatial_ops}
\end{equation}
where $\mathrm{Conv}_{t}^{p}(\cdot)$ and $\mathrm{Conv}_{s}^{p}(\cdot)$ denote convolution along the temporal dimension and across EEG channels, respectively. $\mathrm{BN}(\cdot)$ denotes batch normalization, and $\phi(\cdot)$ denotes a nonlinear activation function.

The TS and ST paths use the same types of encoding operations but apply them in different orders with independently learned parameters. Specifically, the TS path first extracts temporal dynamics and then models cross-channel spatial dependencies, whereas the ST path follows the reverse order:
\begin{equation}
	\begin{aligned}
		H_{\mathrm{TS}}
		&=
		\mathcal{S}^{\mathrm{TS}}
		\left(
		\mathcal{T}^{\mathrm{TS}}(X)
		\right), \\
		H_{\mathrm{ST}}
		&=
		\mathcal{T}^{\mathrm{ST}}
		\left(
		\mathcal{S}^{\mathrm{ST}}(X)
		\right).
	\end{aligned}
	\label{eq:dual_path_encoding}
\end{equation}

\subsubsection{Multiscale Temporal Enhancement and Feature Projection}

\begin{figure}[!t]
	\centering
	\includegraphics[width=\columnwidth]{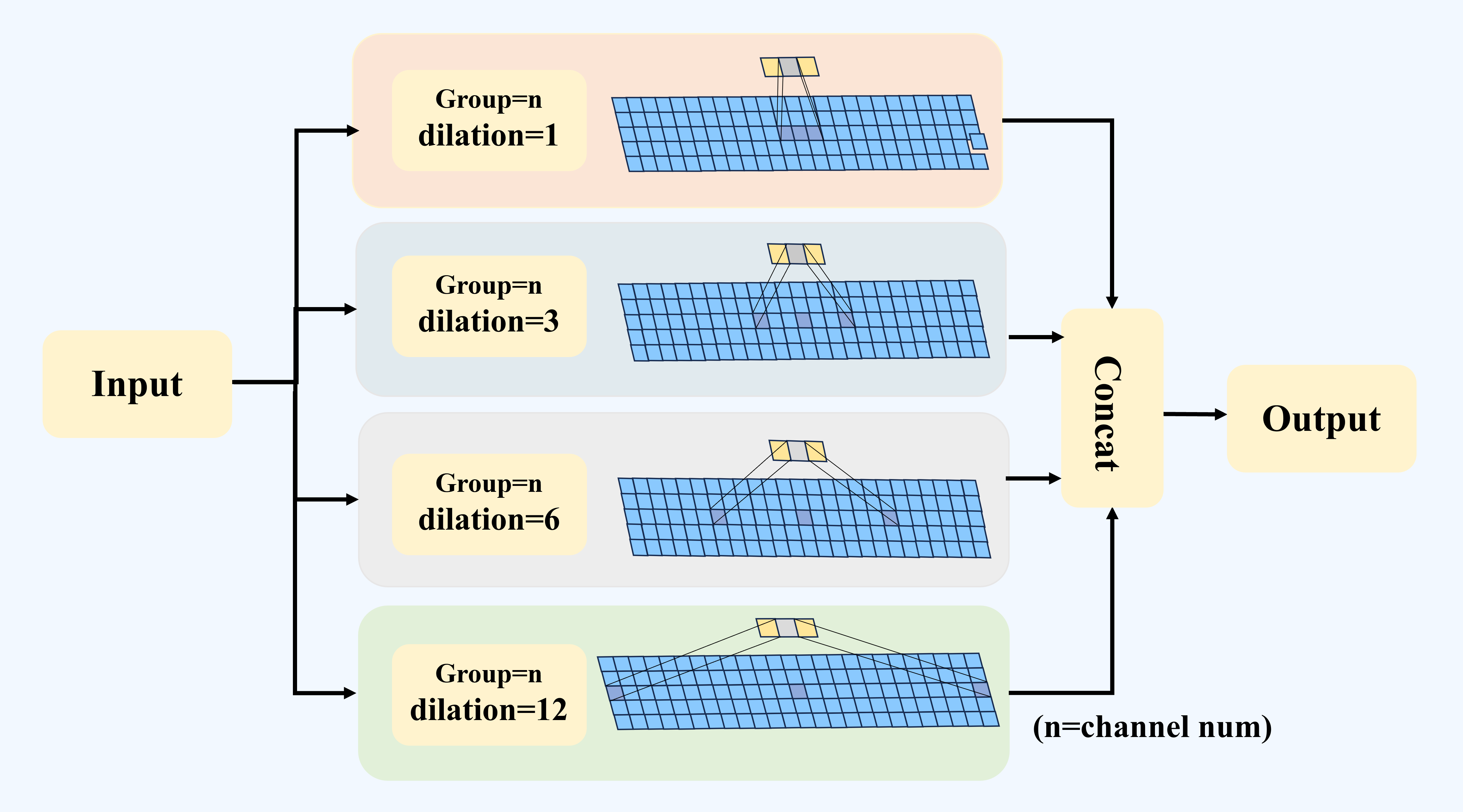}
	\caption{Illustration of the multiscale dilated temporal convolution module. Parallel temporal convolution branches with different dilation rates capture temporal dependencies at multiple receptive-field scales, and their outputs are concatenated to form the enhanced representation.}
	\label{fig:multiscale_temporal}
\end{figure}

To capture emotion-related dynamics at different temporal scales, multiscale dilated temporal convolutions are incorporated into both paths. As illustrated in Fig.~\ref{fig:multiscale_temporal}, parallel temporal convolution branches with different dilation rates capture temporal dependencies over multiple receptive-field scales, and their outputs are concatenated to form an enhanced temporal representation. A temporal attention mechanism then weights temporal positions to emphasize emotion-relevant segments. After pooling, the high-level feature from each path is mapped into a shared representation space through an independent projection head and further $\ell_2$-normalized for subsequent cross-subject contrastive learning and cross-path semantic alignment.

\subsection{Cross-Subject and Cross-Path Collaborative Representation Learning}
\label{subsec:collaborative_learning}

\begin{figure}[!t]
	\centering
	\includegraphics[width=\columnwidth]{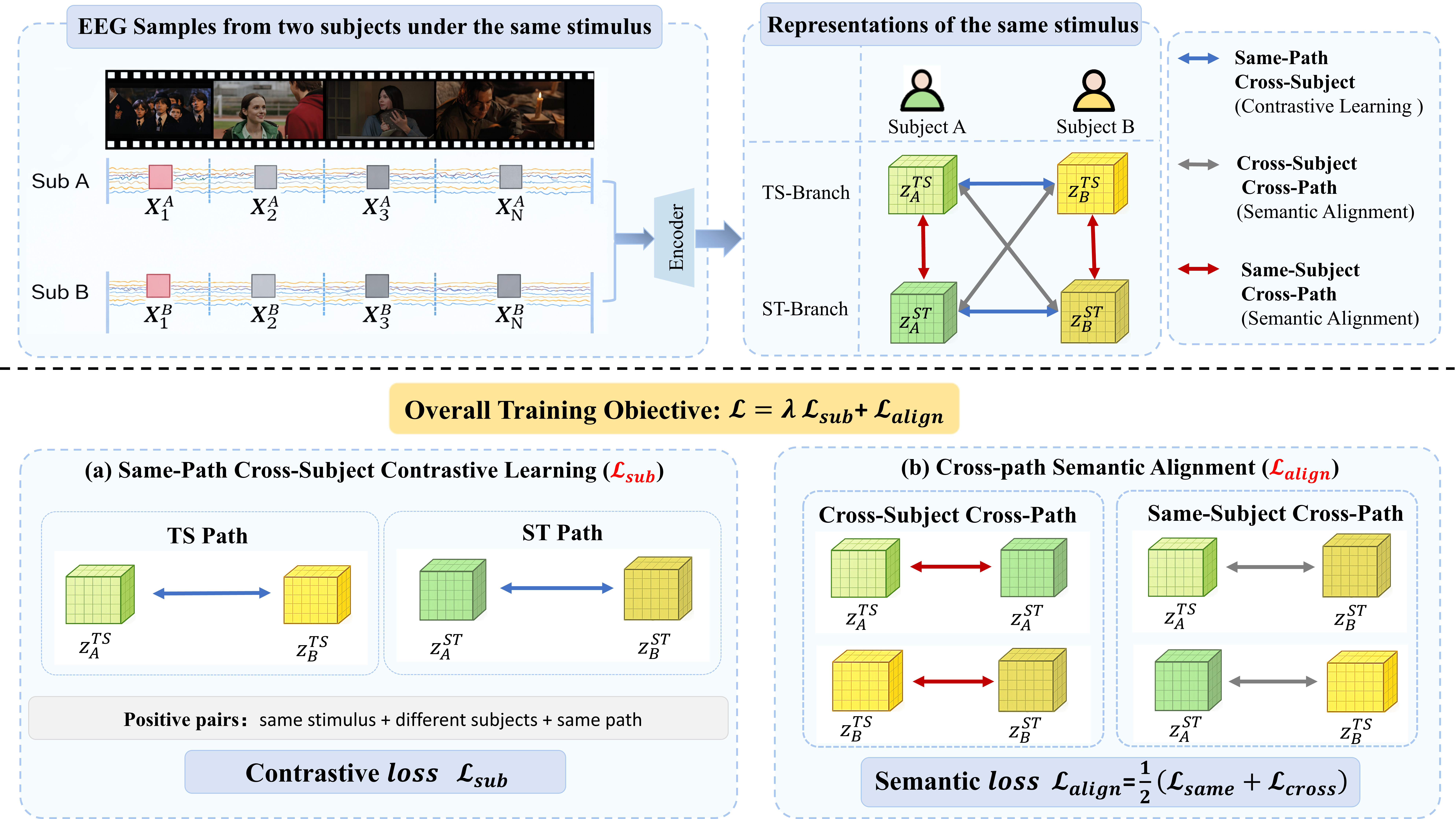}
	\caption{Illustration of the proposed collaborative representation learning scheme, including same-path cross-subject contrastive learning and cross-path semantic alignment.}
	\label{fig:collaborative_learning}
\end{figure}

After dual-path spatiotemporal encoding, the TS and ST representations are jointly optimized through same-path cross-subject contrastive learning and cross-path semantic alignment, as illustrated in Fig.~\ref{fig:collaborative_learning}. The former promotes subject-invariant and stimulus-discriminative representations within each path, whereas the latter aligns the two paths through same-subject and cross-subject cross-path relations.

\subsubsection{Same-Path Cross-Subject Contrastive Learning}

Let a mini-batch contain $N$ EEG samples. For the $n$-th sample, let $z_n^p$ denote the normalized projected representation generated by path $p\in\{\mathrm{TS},\mathrm{ST}\}$. To unify the two paths, each view is represented as $v=(n,p)$, with the corresponding normalized representation $\bar{z}_v=z_n^p$. Each view $v$ is associated with a stimulus label $y_v$, subject label $s_v$, and path label $p_v$. All normalized TS and ST representations in the mini-batch form the representation set $\mathcal{B}$.

Same-path cross-subject contrastive learning aims to learn stimulus-related representations that are robust to subject variation within each path. For an anchor $v$, representations from the same stimulus, different subjects, and the same path are treated as positive samples. The corresponding positive set is defined as
\begin{equation}
	P_{\mathrm{sub}}(v)
	=
	\left\{
	u\neq v
	\mid
	y_u=y_v,\;
	s_u\neq s_v,\;
	p_u=p_v
	\right\}.
	\label{eq:positive_set}
\end{equation}

To restrict contrastive learning to the same path, define the path-specific candidate set as
$\mathcal{B}_p=\{a\in\mathcal{B}\mid p_a=p\}$.
For each positive sample $u\in P_{\mathrm{sub}}(v)$, the pairwise contrastive loss is defined as
\begin{equation}
	\ell(v,u)
	=
	-\log
	\frac{
		\exp\!\left(
		\operatorname{sim}(\bar{z}_v,\bar{z}_u)/\tau
		\right)
	}{
		\displaystyle
		\sum_{\substack{
				a\in\mathcal{B}_{p_v}\\
				a\neq v
		}}
		\exp\!\left(
		\operatorname{sim}(\bar{z}_v,\bar{z}_a)/\tau
		\right)
	}.
	\label{eq:pairwise_contrastive}
\end{equation}
where $\operatorname{sim}(\cdot,\cdot)$ denotes the similarity function and $\tau$ is the temperature parameter. The denominator contains only representations from the same path as the anchor, so the contrastive objective is optimized separately for the TS and ST paths.

Valid anchors in path $p$ are collected as
$V_p=\{v\in\mathcal{B}_p\mid |P_{\mathrm{sub}}(v)|>0\}$.
The losses are first averaged over all cross-subject positive samples for each anchor, then over all valid anchors within each path, and finally over the TS and ST paths:
\begin{equation}
	\mathcal{L}_{\mathrm{sub}}
	=
	\frac{1}{2}
	\sum_{p\in\{\mathrm{TS},\mathrm{ST}\}}
	\frac{1}{|\mathcal{V}_p|}
	\sum_{v\in\mathcal{V}_p}
	\frac{1}{|P_{\mathrm{sub}}(v)|}
	\sum_{u\in P_{\mathrm{sub}}(v)}
	\ell(v,u).
	\label{eq:subject_contrastive}
\end{equation}

This objective reduces subject-specific variation while preserving stimulus-discriminative information within each path.

\subsubsection{Cross-Path Semantic Alignment}

Based on the representations learned through same-path cross-subject contrastive learning, we further align the TS and ST representations under the same stimulus to reduce discrepancies caused by different modeling orders. Unlike the contrastive objective, cross-path semantic alignment introduces no negative samples and directly minimizes the mean squared error (MSE) between paired representations. According to subject identity, cross-path alignment is divided into same-subject and cross-subject relations.

Let $n(v)$ denote the sample index corresponding to view $v$. For same-subject cross-path alignment, the TS and ST representations of the same EEG sample form the pair set
\begin{equation}
	Q_{\mathrm{same}}
	=
	\left\{
	(v,u)
	\mid
	n(v)=n(u),\;
	p_v=\mathrm{TS},\;
	p_u=\mathrm{ST}
	\right\}.
	\label{eq:q_same}
\end{equation}

For cross-subject cross-path alignment, representations from the same stimulus but different subjects and different paths are paired:
\begin{equation}
	\begin{aligned}
		Q_{\mathrm{cross}}
		=
		\big\{(v,u)\mid\;& y_v=y_u,\; s_v\neq s_u,\\
		& p_v=\mathrm{TS},\; p_u=\mathrm{ST}\big\}.
	\end{aligned}
	\label{eq:q_cross}
\end{equation}

The corresponding alignment losses are
\begin{equation}
	\begin{aligned}
		\mathcal{L}_{\mathrm{same}}
		&=
		\frac{1}{|Q_{\mathrm{same}}|}
		\sum_{(v,u)\in Q_{\mathrm{same}}}
		\operatorname{MSE}(\bar{z}_v,\bar{z}_u),
		\\
		\mathcal{L}_{\mathrm{cross}}
		&=
		\frac{1}{|Q_{\mathrm{cross}}|}
		\sum_{(v,u)\in Q_{\mathrm{cross}}}
		\operatorname{MSE}(\bar{z}_v,\bar{z}_u).
	\end{aligned}
	\label{eq:cross_path_losses}
\end{equation}

Since cross-subject pairs are generally more numerous, the two relation types are averaged separately and then equally weighted to avoid pair-count imbalance:
\begin{equation}
	\mathcal{L}_{\mathrm{align}}
	=
	\frac{1}{2}
	\left(
	\mathcal{L}_{\mathrm{same}}
	+
	\mathcal{L}_{\mathrm{cross}}
	\right).
	\label{eq:alignment_loss}
\end{equation}

The final pretraining objective combines same-path cross-subject contrastive learning and cross-path semantic alignment:
\begin{equation}
	\mathcal{L}
	=
	\mathcal{L}_{\mathrm{sub}}
	+
	\lambda\mathcal{L}_{\mathrm{align}},
	\label{eq:overall_objective}
\end{equation}
where $\lambda$ controls the contribution of cross-path semantic alignment. Together, the two objectives reduce subject-specific and modeling-path variations while preserving stimulus-discriminative representations.

\subsection{LoRA-Based Few-Shot Subject Adaptation}

Despite cross-subject pretraining, residual individual differences may remain for unseen subjects. To enable few-shot subject adaptation, we freeze the pretrained encoder and adapt only the downstream classifier in a parameter-efficient manner.

Let $f_{\theta}(\cdot)$ denote the pretrained EEG encoder. After removing the projection heads, its output
$h=f_{\theta}(x)\in\mathbb{R}^{d_{\mathrm{in}}}$
is used as the downstream feature, while the encoder parameters $\theta$ remain frozen during adaptation. For a linear classifier with weight matrix
$W\in\mathbb{R}^{d_{\mathrm{out}}\times d_{\mathrm{in}}}$,
Low-Rank Adaptation (LoRA) introduces a trainable low-rank update:
\begin{equation}
	y
	=
	\left(
	W+\frac{\alpha}{r}BA
	\right)h,
	\label{eq:lora}
\end{equation}
where
$A\in\mathbb{R}^{r\times d_{\mathrm{in}}}$,
$B\in\mathbb{R}^{d_{\mathrm{out}}\times r}$,
and
$r\ll\min(d_{\mathrm{in}},d_{\mathrm{out}})$.
Here, $r$ denotes the low-rank dimension and $\alpha/r$ controls the update scale. During adaptation, $W$ remains frozen and only $A$ and $B$ are optimized, reducing the number and degrees of freedom of trainable parameters under limited labeled data.
Using only a few labeled samples from the target subject, this constrained update adapts the classifier while preserving the pretrained representation space. Overall, DP-STCL separates cross-subject representation learning from parameter-efficient target-subject adaptation, allowing the encoder to preserve shared representations while the low-rank classifier accounts for residual subject-specific variation.

\section{Result}

This section evaluates DP-STCL on four public EEG emotion datasets: SEED~\cite{ref31,ref32},
SEED-IV~\cite{ref57}, SEED-V~\cite{ref58}, and FACED~\cite{ref54}.
We report cross-subject performance comparisons,
ablation studies, few-shot adaptation results, and model efficiency analyses.

\subsection{Dataset}

SEED contains 62-channel EEG recordings from 15 subjects across three sessions. Fifteen movie clips in each session elicit positive, neutral, and negative emotions, yielding a three-class emotion recognition task.

SEED-IV contains EEG recordings from 15 subjects over three sessions. Each session includes 24 video trials covering four emotions: neutral, sadness, fear, and happiness. We therefore evaluate four-class emotion recognition on this dataset.

SEED-V extends SEED to five emotion categories: joy, sadness, disgust, fear, and neutral. It contains 62-channel EEG recordings from 16 subjects, each completing three sessions with 15 movie trials per session. We evaluate five-class emotion recognition on SEED-V.

FACED contains 32-channel EEG recordings from 123 subjects and covers nine emotion categories: anger, fear, disgust, sadness, pleasure, inspiration, joy, tenderness, and neutral. Each subject completed 28 emotion-induction trials. We evaluate both nine-class recognition and binary valence classification, where pleasure, inspiration, joy, and tenderness are treated as positive emotions, and anger, fear, disgust, and sadness as negative emotions.

\subsection{Experimental Setup}

For FACED, we followed the official preprocessing protocol and
downsampled the EEG signals to 250~Hz. For SEED, SEED-IV, and
SEED-V, the signals were downsampled to 125~Hz, bandpass filtered
at 0.5--47~Hz, and segmented into trials according to stimulus
onset and offset.

Noisy channels were detected using two amplitude-duration criteria,
with $(m,n)=(3,0.4)$ for long-duration artifacts and
$(30,0.01)$ for large transient artifacts, where $m$ denotes the
amplitude multiple of the median and $n$ the corresponding duration
proportion. The detected noisy channels were interpolated using their
three nearest neighboring channels. Independent Component Analysis
(ICA) was then applied, and components labeled by EEGLAB as eye- or
muscle-related with confidence above 0.8 were removed.

Sliding-window segmentation used window/step sizes of 5/2~s for FACED,
30/15~s for SEED, 25/9~s for SEED-IV, and 22/11~s for SEED-V,
yielding an average of 13 samples per trial for each dataset.

SEED, SEED-IV, and SEED-V were evaluated using leave-one-subject-out
validation, with one subject used for testing and the remaining subjects
for training in each round. For FACED, 10-fold cross-subject validation
was adopted by dividing the 123 subjects into nine groups of 12 subjects
and one group of 15 subjects. During pretraining, the learning rate and
weight decay were set to $7\times10^{-4}$ and $1.5\times10^{-4}$,
respectively, with 30 training rounds and an early-stopping patience of
10 rounds.

\subsection{Effectiveness of Dual-Path Spatiotemporal Contrastive Pretraining}

To evaluate the effectiveness of DP-STCL, we compare it with the DE+MLP 
baseline and representative cross-subject methods, including 
DSAN~\cite{ref59}, DMMR~\cite{ref60},
MS-MDA~\cite{ref61}, ADAST~\cite{ref62}, GMoE~\cite{ref63},
PGCN~\cite{ref64}, GMSS~\cite{ref65}, CLISA~\cite{ref13},
MSHCL~\cite{ref67}, and DAEST~\cite{ref68}. 

Table~\ref{tab:seed} presents the cross-subject emotion recognition
accuracy of DP-STCL and the competing methods on the SEED, SEED-IV, and
SEED-V datasets. As shown in the table, DP-STCL consistently achieves
higher accuracy than the compared methods across all three datasets.
To further evaluate the generalizability of DP-STCL, we conducted binary
classification (cls2) and nine-class classification (cls9) experiments on
the FACED dataset, with the results reported in Table~\ref{tab:faced}.
DP-STCL also achieved strong performance on FACED. 

\begin{table}[!t]
	\caption{Cross-Subject Emotion Recognition Performance on the SEED,
		SEED-IV, and SEED-V Datasets. Results marked with $^\dagger$ are
		reproduced; others are from original papers.}
	\label{tab:seed}
	\centering
	\renewcommand{\arraystretch}{1.2}
	\setlength{\tabcolsep}{4pt}
	\begin{tabular}{lccc}
		\toprule
		\multicolumn{1}{c}{\textbf{Method}} &
		\textbf{SEED (Acc\%)} &
		\textbf{SEED-IV (Acc\%)} &
		\textbf{SEED-V (Acc\%)} \\
		\midrule
		DE+MLP           & $79.9\pm8.7$           & $59.3\pm17.2$           & $59.3\pm17.2$ \\
		DSAN~\cite{ref59}             & $84.1\pm7.6$           & $62.2\pm14.2$           & $59.3\pm16.8$ \\
		DMMR~\cite{ref60}             & $88.3\pm5.6$           & $72.7\pm8.0$            & /             \\
		MS-MDA~\cite{ref61}           & $84.5\pm7.5$           & $64.3\pm11.1$           & $58.2\pm12.3$ \\
		ADAST~\cite{ref62}            & $87.3\pm2.1$           & $53.7\pm13.6$           & $50.7\pm14.0$ \\
		GMoE~\cite{ref63}             & $84.6\pm9.3$           & $71.2\pm8.5$            & $76.3\pm9.2$  \\
		PGCN~\cite{ref64}             & $84.6\pm8.7$           & $73.7\pm7.2$            & $65.2\pm10.1$ \\
		GMSS~\cite{ref65}             & $86.3\pm6.9$           & $62.6\pm10.3$           & /             \\
		CLISA$^\dagger$~\cite{ref13}  & $86.4\pm6.4$           & /                       & $67.3\pm13.0$ \\
		DAEST$^\dagger$~\cite{ref68}  & $87.7\pm3.6$           & $73.8\pm8.7$            & $70.2\pm12.5$ \\
		\midrule
		\multicolumn{1}{c}{\textbf{DP-STCL}} & $\mathbf{90.3\pm10.2}$ & $\mathbf{76.8\pm10.4}$ & $\mathbf{79.0\pm11.4}$ \\
		\bottomrule
	\end{tabular}
\end{table}

\begin{table}[!t]
	\caption{Cross-Subject Emotion Recognition Performance on the FACED Dataset}
	\label{tab:faced}
	\centering
	\renewcommand{\arraystretch}{1.2}
	\setlength{\tabcolsep}{3pt}
	\begin{tabular}{l>{\centering\arraybackslash}p{1.4cm}>{\centering\arraybackslash}p{1.4cm}>{\raggedright\arraybackslash}p{3.8cm}}
		\toprule
		\multicolumn{1}{c}{\textbf{Method}} &
		\makecell{\textbf{FACED} \\ \textbf{(cls2)} \\ \textbf{(Acc\%)}} &
		\makecell{\textbf{FACED} \\ \textbf{(cls9)} \\ \textbf{(Acc\%)}} &
		\multicolumn{1}{c}{\textbf{Source of Result}} \\
		\midrule
		DE+MLP  & $60.9\pm3.2$ & $21.7\pm2.2$ & Baseline \\
		SimCLR  & /            & $30.2\pm2.7$ & Reproduced Results in~\cite{ref20} \\
		MSHCL   & $77.0\pm3.6$ & $45.7\pm3.4$ & Original Results in~\cite{ref67} \\
		CLISA   & $67.8\pm4.1$ & $42.4\pm3.1$ & Reproduced Results in~\cite{ref13} \\
		DAEST   & $77.8\pm5.8$ & $62.3\pm8.1$ & Reproduced Results in~\cite{ref68} \\
		\midrule
		\multicolumn{1}{c}{\textbf{DP-STCL}} & $\mathbf{80.5\pm4.4}$ & $\mathbf{66.5\pm7.8}$ & Results from this paper \\
		\bottomrule
	\end{tabular}
\end{table}

\begin{figure}[!t]
	\centering
	\includegraphics[width=\columnwidth]{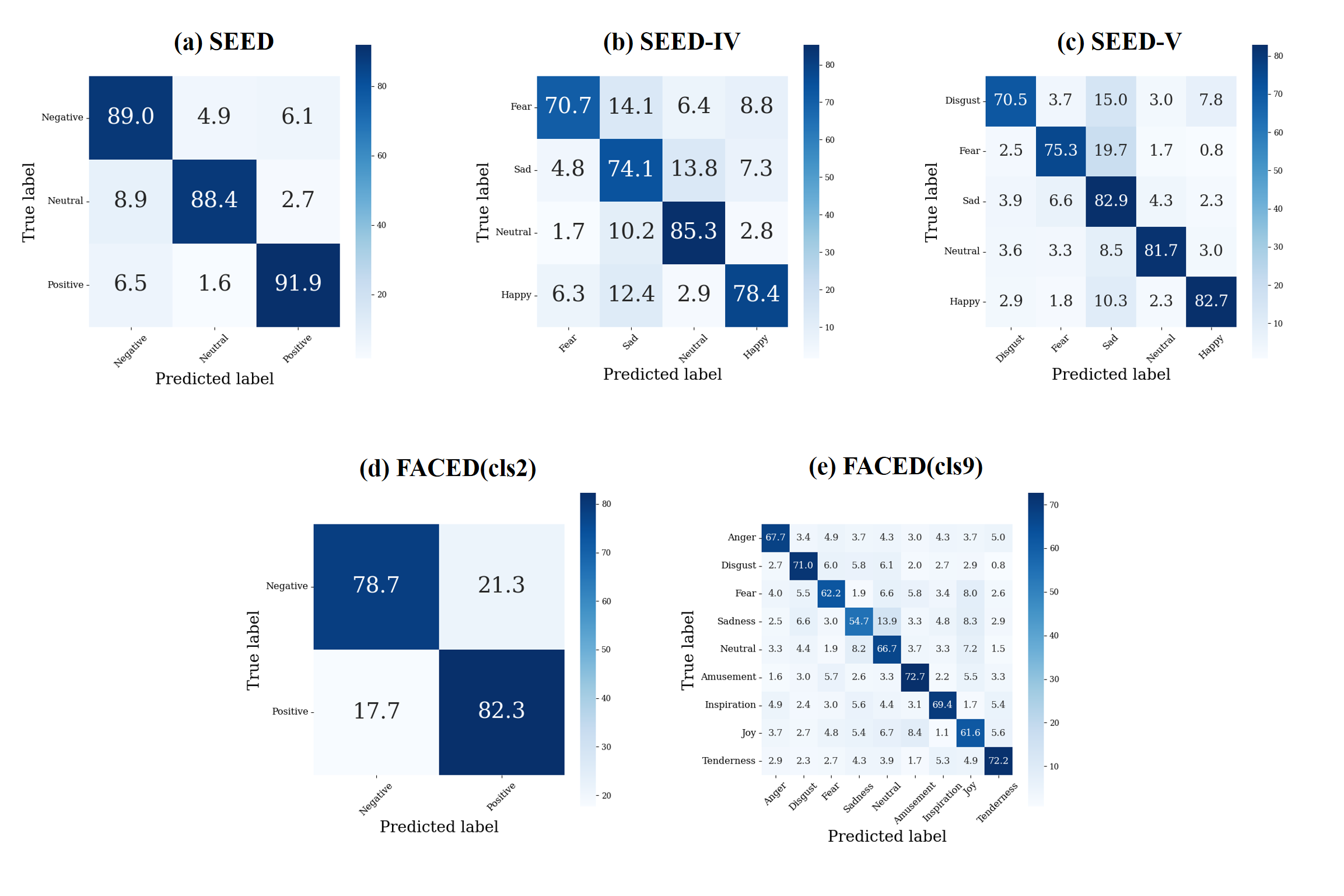}
	\caption{Confusion matrices for emotion classification on the SEED,
		SEED-IV, SEED-V, and FACED datasets.}
	\label{fig:confusion}
\end{figure}

To provide a more intuitive visualization of DP-STCL's classification
performance across different emotion categories, we plotted confusion
matrices for the SEED, SEED-IV, SEED-V, and FACED datasets, as shown in
Fig.~\ref{fig:confusion}.

\subsection{Effectiveness of Dual-Path Spatiotemporal Modeling}

To evaluate the dual-path architecture and joint training mechanism,
we compare single-path independent training with DP-STCL dual-path
joint training, as reported in Table~\ref{tab:ablation_path}. In the former,
the TS-only and ST-only models are trained separately using same-path
cross-subject contrastive learning. In the latter, the two paths are
jointly optimized with cross-path semantic alignment, and the individual TS and ST results are obtained by evaluating each branch
of the jointly trained model rather than by separately retraining the
two branches.

\begin{figure*}[!t]
	\centering
	\includegraphics[width=\textwidth]{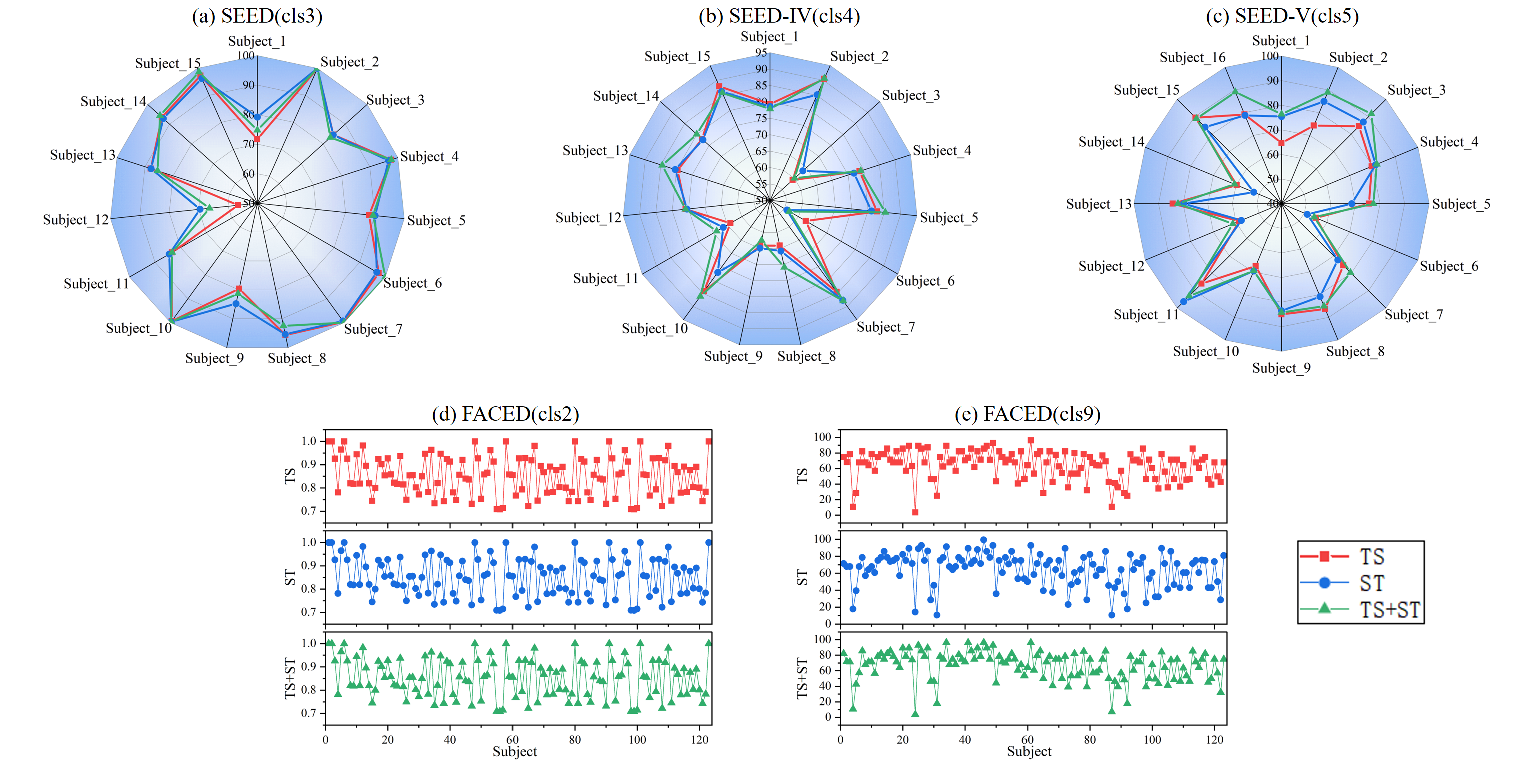}
	\caption{Subject-wise accuracy distribution of dual-path
		spatiotemporal modeling.}
	\label{fig:subject_compare}
\end{figure*}

Both independently trained paths achieve competitive performance, while
their relative advantages vary across datasets, supporting the
complementarity of the two spatiotemporal modeling orders. Joint training
further improves both individual branches across all five tasks. Compared
with independent training, the jointly trained ST branch improves by
0.6--1.6 percentage points, while the TS branch improves by 1.0--2.2
points. This indicates that the benefit of DP-STCL is not solely due to
feature fusion at inference, but also arises from improved branch
representations during collaborative training. Fig.~\ref{fig:subject_compare}
further shows the subject-wise accuracy distributions of the jointly
trained TS, ST, and fused representations.

The fused TS+ST representation achieves the highest accuracy across
all five tasks in Table~\ref{tab:ablation_path}. Relative to the stronger
jointly trained single branch, fusion yields gains of 0.8, 1.1, 3.7,
0.7, and 2.4 percentage points on SEED, SEED-IV, SEED-V, FACED cls2,
and FACED cls9, respectively. The larger gains on SEED-V and FACED
cls9 further suggest that the two paths retain complementary discriminative
information despite cross-path semantic alignment.

\begin{table*}[!t]
	\caption{Ablation Results for Dual-Path Spatiotemporal Modeling and the Joint Training Mechanism}
	\label{tab:ablation_path}
	\centering
	\renewcommand{\arraystretch}{1.2}
	\begin{tabular}{llccccc}
		\toprule
		\multicolumn{1}{c}{\textbf{Training Strategy}} &
		\multicolumn{1}{c}{\textbf{Test Repr.}} &
		\textbf{SEED (Acc\%)} &
		\textbf{SEED-IV (Acc\%)} &
		\textbf{SEED-V (Acc\%)} &
		\textbf{FACED cls2 (Acc\%)} &
		\textbf{FACED cls9 (Acc\%)} \\
		\midrule
		\multirow{2}{*}{Single-Path Independent Training}
		& ST-only  & $88.9\pm9.2$  & $73.5\pm9.5$  & $73.6\pm12.5$ & $78.5\pm4.5$ & $63.0\pm7.5$ \\
		& TS-only  & $87.1\pm9.4$  & $74.3\pm9.9$  & $74.2\pm10.8$ & $78.0\pm4.2$ & $62.5\pm7.0$ \\
		\midrule
		\multirow{3}{*}{DP-STCL Dual-Path Joint Training}
		& ST       & $89.5\pm10.5$ & $74.9\pm9.1$  & $75.2\pm12.3$ & $79.8\pm4.5$ & $64.1\pm7.4$ \\
		& TS       & $89.3\pm11.0$ & $75.7\pm9.9$  & $75.3\pm10.7$ & $79.0\pm4.5$ & $64.0\pm7.0$ \\
		& \textbf{TS+ST} & $\mathbf{90.3\pm10.2}$ & $\mathbf{76.8\pm10.4}$ & $\mathbf{79.0\pm11.4}$ & $\mathbf{80.5\pm4.4}$ & $\mathbf{66.5\pm7.8}$ \\
		\bottomrule
	\end{tabular}
\end{table*}

\subsection{Analysis of the Cross-Path Semantic Alignment Strategy}

We further analyze the cross-path semantic alignment strategy by
examining the choice of cross-path learning objective and the weight
of semantic alignment.

\begin{figure}[!t]
	\centering
	\includegraphics[width=\columnwidth]{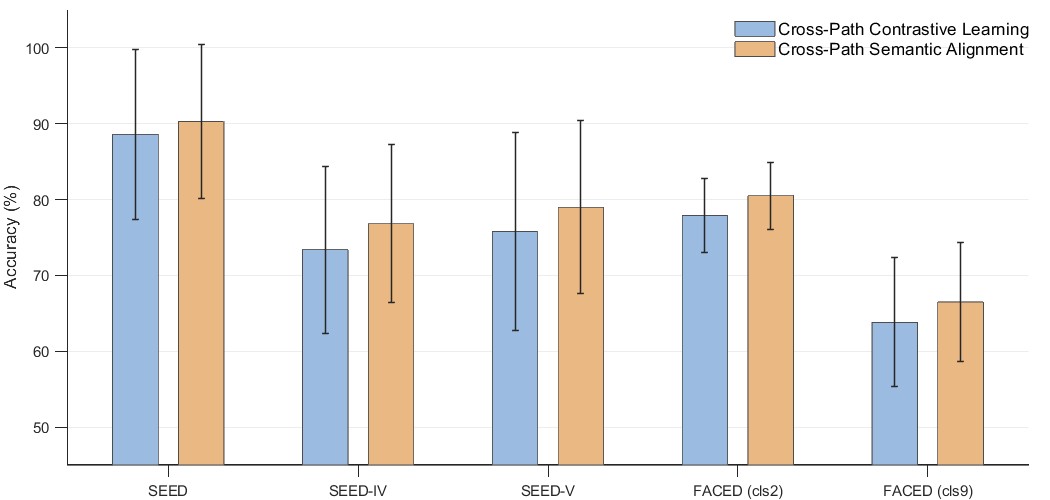}
	\caption{Performance comparison of different cross-path learning
		strategies across datasets.}
	\label{fig:crosspath_strategy}
\end{figure}

We first compare two cross-path learning objectives while keeping
same-path cross-subject contrastive learning unchanged. The first
applies a SimCLR-style contrastive objective to TS--ST representations,
whereas the proposed strategy uses MSE-based semantic alignment for
same-stimulus cross-path pairs. All other network and training
configurations are kept identical.
As shown in Fig.~\ref{fig:crosspath_strategy}, MSE-based semantic
alignment consistently outperforms cross-path contrastive learning
across the evaluated datasets. Since $\mathcal{L}_{\mathrm{sub}}$
already provides stimulus discrimination within each path, introducing
additional positive-negative competition across TS and ST may impose
overly strong constraints on their structurally different
representations. In contrast, $\mathcal{L}_{\mathrm{align}}$ aligns
same-stimulus cross-path representations without introducing
cross-path negative samples. This allows cross-path learning to reduce
representation discrepancies caused by different modeling orders while
preserving the complementary information captured by the two paths.

\begin{figure}[!t]
	\centering
	\includegraphics[width=\columnwidth]{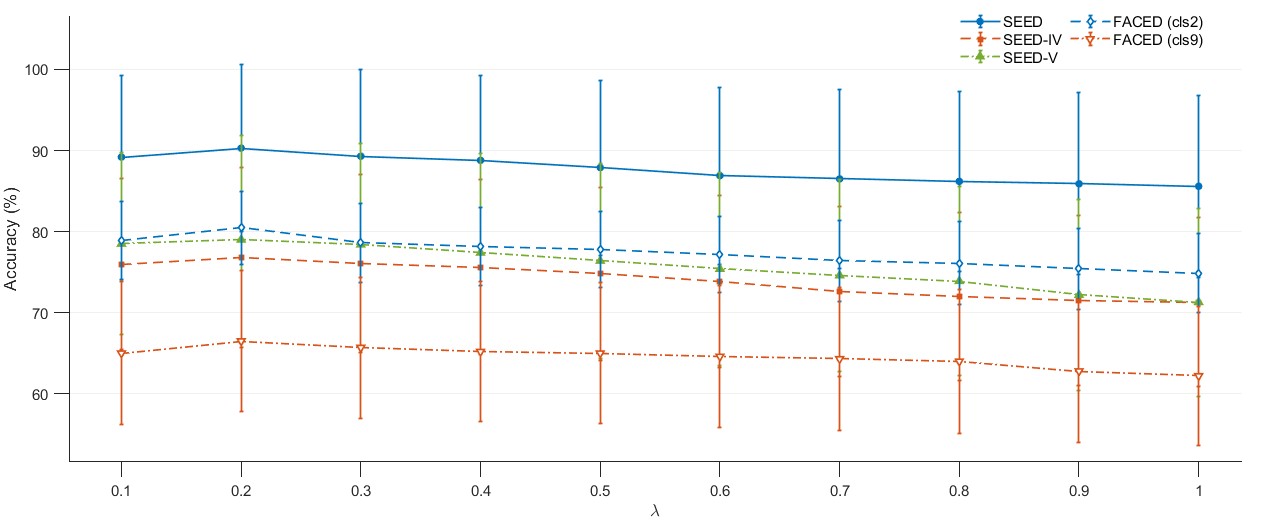}
	\caption{Sensitivity analysis of the cross-path semantic alignment
		weight $\lambda$ across datasets.}
	\label{fig:lambda}
\end{figure}

We next examine the sensitivity to the semantic alignment weight
$\lambda$, which controls the contribution of
$\mathcal{L}_{\mathrm{align}}$ during pretraining. As shown in
Fig.~\ref{fig:lambda}, $\lambda=0.2$ yields stable and
relatively strong performance across the evaluated datasets. When
$\lambda=0$, training relies solely on same-path cross-subject
contrastive learning, leaving no explicit cross-path constraint.
A moderate increase in $\lambda$ reduces representation discrepancies
between the two paths and promotes collaborative learning. However,
an excessively large $\lambda$ may overemphasize path consistency and
compress discriminative differences between the TS and ST
representations, thereby weakening their complementarity. These results
indicate that $\mathcal{L}_{\mathrm{sub}}$ serves as the primary
discriminative objective, while $\mathcal{L}_{\mathrm{align}}$ acts as
a complementary semantic bridge between the two paths.

\subsection{Few-Shot Performance Analysis}

We evaluate few-shot adaptation under 3-shot, 5-shot, and 10-shot
settings by comparing Full-finetune, Head-only, and LoRA on SEED,
SEED-IV, SEED-V, FACED (cls2), and FACED (cls9). The classification
results are reported in Table~\ref{tab:fewshot}.

Full-finetune updates the entire network using labeled target-subject
samples. Head-only keeps the feature encoder fixed and updates the
classification head, whereas LoRA freezes the pretrained parameters
and optimizes only the low-rank updates introduced into the classifier.

As shown in Table~\ref{tab:fewshot}, Full-finetune achieves the
highest classification accuracy across all evaluated datasets and shot
settings, reflecting the greater adaptation flexibility obtained by
updating the entire network. LoRA nevertheless remains close to
Full-finetune while requiring substantially fewer trainable parameters.
In particular, relatively small performance gaps are observed on SEED,
SEED-IV, and FACED across different few-shot settings, indicating that
low-rank adaptation can effectively accommodate residual
subject-specific variation without modifying the pretrained encoder.

LoRA also consistently outperforms Head-only across the evaluated
datasets and shot settings. Since both strategies keep the pretrained
encoder fixed, this improvement is more reasonably attributed to the
constrained low-rank update rather than greater adaptation freedom.
By restricting classifier adaptation to a compact low-rank subspace,
LoRA can reduce the risk of overfitting under limited labeled data
while preserving the pretrained representation. Overall, LoRA achieves
competitive adaptation performance with substantially lower
parameter-update overhead than Full-finetune, providing a favorable
balance between target-subject adaptation and parameter efficiency in
few-shot cross-subject EEG emotion recognition.

\begin{table}[!t]
	\caption{Comparison of Classification Accuracy under Different
		Few-Shot Settings and Parameter Update Strategies}
	\label{tab:fewshot}
	\centering
	\renewcommand{\arraystretch}{1.2}
	\setlength{\tabcolsep}{3pt}
	\scriptsize
	\begin{tabular}{llccc}
		\toprule
		\multicolumn{1}{c}{\textbf{Dataset}} &
		\multicolumn{1}{c}{\textbf{Method}} &
		\makecell{\textbf{3-shot} \\ \textbf{(Acc\%)}} &
		\makecell{\textbf{5-shot} \\ \textbf{(Acc\%)}} &
		\makecell{\textbf{10-shot} \\ \textbf{(Acc\%)}} \\
		\midrule
		\multirow{3}{*}{SEED}
		& LoRA          & $91.1\pm7.9$  & $91.2\pm7.9$  & $91.3\pm7.5$  \\
		& Head-only     & $77.8\pm10.4$ & $79.8\pm9.2$  & $79.9\pm10.7$ \\
		& Full-finetune & $91.6\pm7.2$  & $92.8\pm6.2$  & $94.3\pm5.1$  \\
		\midrule
		\multirow{3}{*}{SEED-IV}
		& LoRA          & $77.2\pm9.6$  & $77.9\pm8.4$  & $83.6\pm5.5$  \\
		& Head-only     & $63.1\pm9.5$  & $60.8\pm10.8$ & $62.7\pm10.1$ \\
		& Full-finetune & $77.5\pm9.3$  & $78.6\pm8.4$  & $85.8\pm5.9$  \\
		\midrule
		\multirow{3}{*}{SEED-V}
		& LoRA          & $79.2\pm11.0$ & $80.1\pm11.3$ & $83.9\pm11.0$ \\
		& Head-only     & $60.4\pm12.3$ & $59.7\pm12.1$ & $61.5\pm11.6$ \\
		& Full-finetune & $80.3\pm10.8$ & $81.0\pm10.6$ & $84.7\pm10.0$ \\
		\midrule
		\multirow{3}{*}{FACED (cls2)}
		& LoRA          & $82.8\pm9.9$  & $87.4\pm8.7$  & $92.9\pm5.7$  \\
		& Head-only     & $65.2\pm10.9$ & $64.8\pm11.8$ & $64.8\pm10.4$ \\
		& Full-finetune & $84.2\pm9.1$  & $88.3\pm7.8$  & $94.1\pm5.3$  \\
		\midrule
		\multirow{3}{*}{FACED (cls9)}
		& LoRA          & $80.4\pm10.8$ & $86.0\pm14.4$ & $91.8\pm11.1$ \\
		& Head-only     & $36.4\pm12.2$ & $36.0\pm13.7$ & $36.1\pm12.1$ \\
		& Full-finetune & $82.6\pm10.3$ & $87.4\pm14.6$ & $93.0\pm10.5$ \\
		\bottomrule
	\end{tabular}
\end{table}

\subsection{t-SNE Visualization of Learned Representations}

\begin{figure*}[!t]
	\centering
	\includegraphics[width=\textwidth, height=0.40\textheight, keepaspectratio]{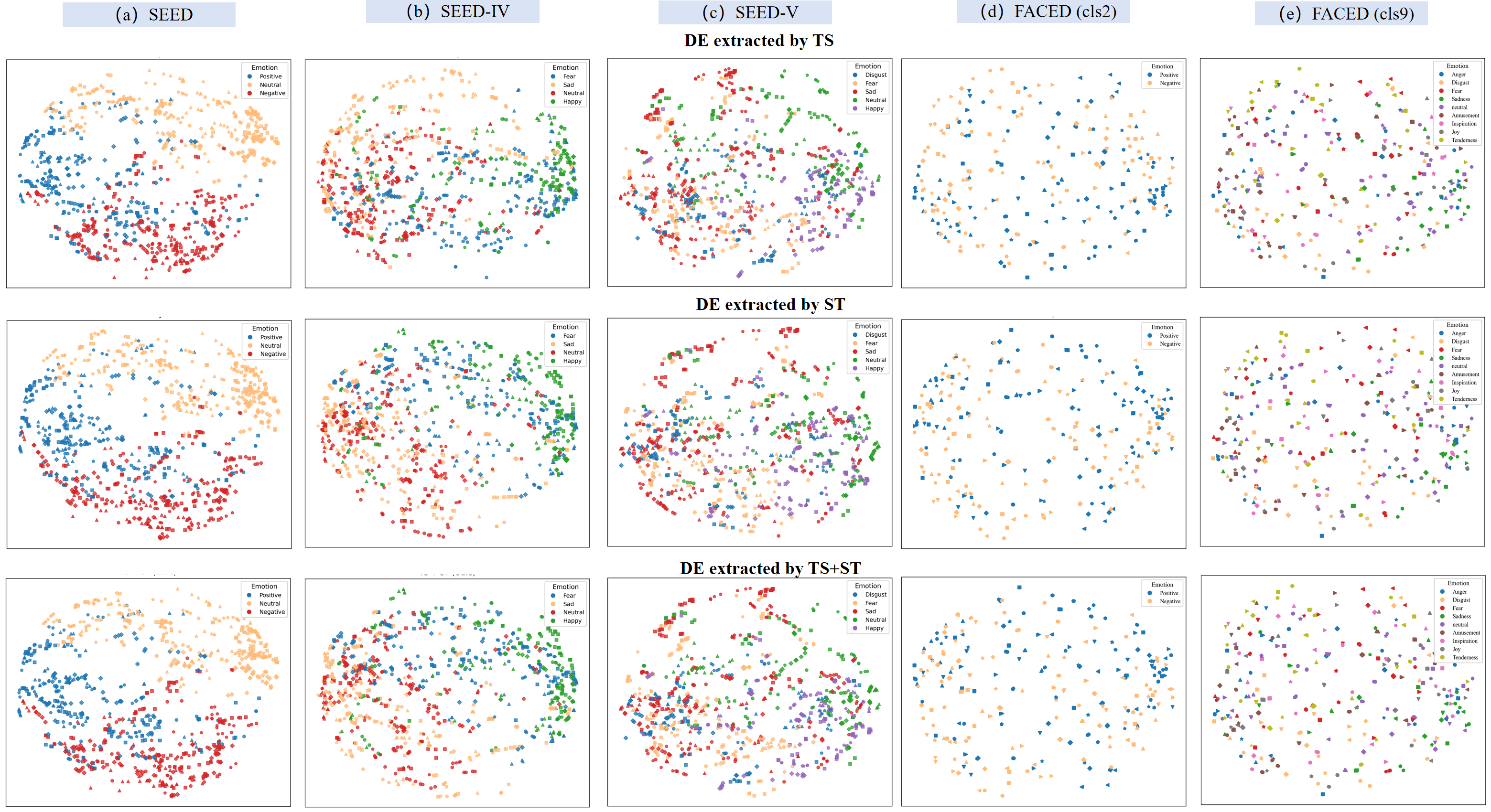}
	\caption{t-SNE visualization of the representations extracted by the
		$\text{TS}$ branch, $\text{ST}$ branch, and fused
		$\text{TS}+\text{ST}$ representation on the five emotion recognition
		tasks.}
	\label{fig:tsne}
\end{figure*}

To visually analyze the discriminability of different feature
representations for emotion classification and their cross-subject
consistency, we visualize the features extracted from the $\text{TS}$
branch, the $\text{ST}$ branch, and their fused $\text{TS}+\text{ST}$
representation using t-SNE, as shown in Fig.~\ref{fig:tsne}. The three
rows correspond to the $\text{TS}$, $\text{ST}$, and
$\text{TS}+\text{ST}$ representations, respectively, while the columns
present the results on SEED, SEED-IV, SEED-V, FACED~(cls2), and
FACED~(cls9). The visualization provides further evidence for the
effectiveness of the proposed dual-path spatiotemporal modeling strategy.
By jointly modeling Temporal-to-Spatial and Spatial-to-Temporal
interactions, the model learns more stable and discriminative
emotion-related EEG representations, which contributes to improved
classification performance.

\section{Conclusion}

This paper proposes DP-STCL for cross-subject EEG emotion recognition
to address inter-subject variability and the limitations of fixed
spatiotemporal modeling orders. The framework employs complementary
Temporal-to-Spatial and Spatial-to-Temporal encoding paths. During
representation learning, same-path cross-subject relations reduce
subject-specific variation, while same-subject and cross-subject
cross-path relations promote semantic consistency across different
modeling orders. Their joint optimization enables the two paths to
preserve complementary characteristics while converging toward a
shared high-level semantic space. For downstream adaptation, the
pretrained encoder is frozen and LoRA is introduced into the classifier,
enabling parameter-efficient adaptation to unseen subjects with limited
labeled data.

Experiments on SEED, SEED-IV, SEED-V, and FACED demonstrate that
DP-STCL achieves strong recognition performance across different
cross-subject emotion classification settings. The results further
support the effectiveness of stimulus-based cross-subject associations,
collaborative dual-path learning, and low-rank few-shot adaptation for
learning stable and transferable EEG representations. Future work will
incorporate brain-region topology into the dual-path architecture,
investigate generalization to unseen stimuli, and extend the framework
to cross-device, cross-session, and multimodal physiological signal
scenarios.
\bibliographystyle{IEEEtran}
\bibliography{references_verified_plus5}

\end{document}